\documentclass[10.5pt,twocolumn,a4paper]{article}
\usepackage[utf8]{inputenc}
\usepackage{amsmath}
\usepackage{amsfonts}
\usepackage{amssymb}
\usepackage{multicol}
\usepackage{graphicx}
\usepackage{float}
\usepackage[none]{hyphenat}
\usepackage[framemethod=tikz]{mdframed}

\usepackage[left=2cm,right=2cm,top=2cm,bottom=2cm]{geometry}

\usepackage{braket}
\usepackage{multirow}

\begin{document}

\newgeometry{left=2cm, right=2cm, top=1.5cm, bottom=2cm}

\title{{\LARGE Batch Normalization and the impact of batch structure on the behavior of deep convolution networks}}
\author{Mohamed  Hajaj  \hspace{1.5cm}   Duncan Gillies\\{\small Department of Computing, Imperial College London}}

\date{}
\maketitle

\newcommand*{\bfrac}[2]{\genfrac{}{}{0pt}{}{#1}{#2}}

\section*{Abstract}
\textit{Batch normalization was introduced in 2015 to speed up training of deep convolution networks by normalizing the activations across the current batch to have zero mean and unity variance. The results presented here show an interesting aspect of batch normalization, where controlling the shape of the training batches can influence what the network will learn. If training batches are structured as balanced batches (one image per class), and inference is also carried out on balanced test batches, using the batch's own means and variances, then the conditional results will improve considerably. The network uses the strong information about easy images in a balanced batch, and propagates it through the shared means and variances to help decide the identity of harder images on the same batch.  Balancing the test batches requires the labels of the test images, which are not available in practice, however further investigation can be done using batch structures that are less strict and might not require the test image labels. The conditional results show the error rate almost reduced to zero for nontrivial datasets with small number of classes such as the CIFAR10.}

\section{Introduction}

When training deep convolution networks, batch normalization BN \cite{ioffe2015batch} reduces the dependency of the gradients on the scale of the parameters and their initial values, which allows for much higher learning rates, much faster convergence, and better final results. The experiments presented here show that by the nature of its implementation, batch normalization has more to it than just speeding up training. Because activations are normalized using means and variances that are shared across all images in the current batch, the output activations of one image are influenced by all the other images in the batch. This means that the behavior of the network is not only receptive to the individual images but also to the structure of the training batches. Therefore, with BN there is an extra level of control that can be used to guide the training of these networks, which is based on how batches are constructed. The results presented here show that it is possible to make the network learn something based on how batches are constructed.

Balanced batches are batches that have a single instance from each class with size equal to the number of classes. If the network is trained only on balanced batches, then in addition to learning how to classify single images, BN will allow the network to learn an extra logic based on the structure of balanced batches. Because it was only exposed to balanced batches in the training phase, the network will learn an association mechanism between the images in the batch through the shared means and variances of BN to always expect balanced batches. If the performance of the network is measured in the standard way on single test images, then this association mechanism based on batch structure cannot be noticed. In order to measure it, the network needs to be tested on balanced test batches using each batch's own means and variances. The practical difficulty here is balancing the test batches, which requires the labels of the test images. Therefore, the purpose of these experiments is to investigate this extra dimension of control based on \textit{batch structure} that is made possible through BN, and it is not claiming state of the art results. However,  because of the scale of improvement (conditional) achieved on the CIFAR10 dataset, the subject is worth further investigation maybe using other batch structures that do not require the test image labels.

\section{Implementation}
\subsection{Batch Normalization}

For a deep neural network, the input distribution of layer $l$ depends on the parameters of all previous layers, and as these parameters change, the input distribution of layer $l$ will also change. Layer $l$ will try to adapt to an input distribution that keeps changing throughout training, and that slows up training. This problem is called internal covariance shifts \cite{shimodaira2000improving}, and BN tries to reduce this problem by performing a simplified version of complete whitening to the inputs of each layer. First, BN assumes that input features are independent, and therefore can be normalized independently to have zero mean and unity variance. Second, the means and variances are calculated across the current batch, and not over the entire training data (hence the name batch normalization). 

\restoregeometry

Because the networks used in this study \cite{he2016deep} are fully convolutional, BN will only be applied to convolution layers. For each output channel a mean and variance are calculated across all images in the current batch. If the channel size is $h\times w$, and the batch size is $m$ images, and $m’ = m\times h\times w$, then the mean and variance are calculated using equations (1) and (2), and each location $x_i$ in that output channel across all the images in the batch is normalized using equation (3). A convolution layer with $N$ output channels will need $N$ pairs of means and variances. The algorithm also adds a linear transformation $y=\alpha x+\beta$ ($\alpha, \beta$ are trainable parameters) after the normalization step to restore the expressive capabilities of the network. For the backward error propagation equations refer to \cite{ioffe2015batch}.

\begin{align}
\mu &= \frac{1}{m'} \sum_{i=1}^{m'} x_i\\
\sigma^2 &=  -\mu^2 + \frac{1}{m'}  \sum_{i=1}^{m'} x_i^2 \\
\hat{x_i}&=\frac{x_i-\mu}{\sqrt{\sigma^2+\varepsilon}}
\end{align}

\subsection{Balanced Batches }
A balanced batch contains a single instance from each class, and therefore the batch size of a balanced batch will always be equal to the number of classes. With balanced batches, the standard BN algorithm will be used with the following considerations at the training and inference stages. 

\begin{itemize}
 \setlength\itemsep{0.1em}
\item \textbf{Training}: - instead of being constructed randomly, training batches are created to be balanced, and to contain a single instance from each class. If training images were shuffled to prevent an image from always appearing in the same batch (to improve performance), then the shuffling subroutine needs to be changed to always generate balanced batches.
\item \textbf{Inference}: - Standard BN calculates a fixed set of means and variances over the entire training data to be used at the inference stage instead of using the means and variances of the current test batch. This is done to make the prediction of the network deterministic and depends only on the image itself, and not on the other images in the batch. In order to measure the effect of training the network on balanced batches, test images are arranged as balanced batches and the current means and variances of the test batch itself are used in the inference process.
\end{itemize}
%\vspace{-0.4cm}
\subsection{Network Structure }
\begin{figure}[H]
\centering
\includegraphics[scale=0.8]{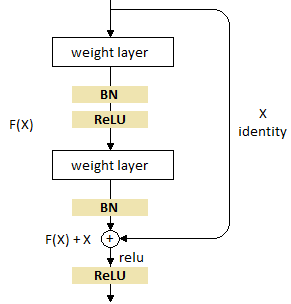}
\caption{{\small Residual Block}} 
\label{Residual_Block}
\end{figure}

Deep residual convolution networks \cite{he2016deep} were used in this study, where a standard deep residual network starts with a single convolution layer, followed by multiple residual building blocks, followed by one fully connected layer, which is the output layer. A residual block figure(\ref{Residual_Block}) is made of 2 or 3 stacked convolution layers warped by identity mapping so that the output of the block is the combination of the input signal to that block and the output signal of the stacked convolution layers. Residual learning makes it possible to train very deep networks with up to hundreds of layers by eliminating the degradation problem \cite{he2015convolutional, srivastava2015highway} exhibited by standard very deep stacked networks. 

%=================================================================

%================================================================

\section{Experiments and Results}
\subsection{CIFAR10 Experiment}

The first experiments to measure the effect of using balanced batches with BN were done using the CIFAR10 dataset. Table (\ref{Table_CIFAR10_NetworkStructure}) shows the structure of the two deep residual networks with depths equal to 20 and 44 layers used in this experiment, and they are very similar to the ones used by He et al. \cite{he2016deep} for CIFAR10. The same treatment of the CIFAR10 dataset by enlarging the images with zero-padding from $32\times32$ to $40\times40$ pixels is used here. The main stream approach in dealing with the CIFAR10 dataset uses cropping without scaling because the images are very small, $32\times32$ pixels. However, we found that using scaling has improved the results despite the images being very small. The weight initialization method in \cite{he2015delving}, and the standard color augmentation method in \cite{krizhevsky2012imagenet} are used. Simple gradient descent with weight decay is used to update the network parameters.

%=================================================================
	
\begin{table}[H]
\centering
\renewcommand{\arraystretch}{1.6}
\begin{tabular}{|c|c|c|}	
	\hline 
	
	$\mathrm{\bfrac{output}{size}}$ & 20 Layers & 44 Layers \\ 
	
	\hline 
	
	\multirow{2}{*}{\scriptsize {$36\times36$}} & \multicolumn{2}{|c|}     {{\footnotesize $conv,\ 3\times3,\hspace{.2cm} 64$}}\\
	\cline{2-3}
	                              & {\small $\left[\bfrac{conv,\ 3\times3,\hspace{.2cm} 64}{conv,\ 3\times3,\hspace{.2cm} 64}\right]\times3$} & {\small $\left[\bfrac{conv,\ 3\times3,\hspace{.2cm} 64}{conv,\ 3\times3,\hspace{.2cm} 64}\right]\times7$}\\[0.1cm]                             
	 
	\hline
	 
	 \multirow{2}{*}{\scriptsize {$18\times18$}} & \multicolumn{2}{|c|}     {{\footnotesize $max\ pool\ 3\times3$, stride 2}}\\
	\cline{2-3}
	                              & {\small $\left[\bfrac{conv,\ 3\times3,\hspace{.2cm} 128}{conv,\ 3\times3,\hspace{.2cm} 128}\right]\times3$} & {\small $\left[\bfrac{conv,\ 3\times3,\hspace{.2cm} 128}{conv,\ 3\times3,\hspace{.2cm} 128}\right]\times7$}\\[0.1cm]                                                          
	
	\hline
	       
    \multirow{2}{*}{\scriptsize {$9\times9$}} & \multicolumn{2}{|c|}     {{\footnotesize $max\ pool\ 3\times3$, stride 2}}\\
	\cline{2-3}
	                              & {\small $\left[\bfrac{conv,\ 3\times3,\hspace{.2cm} 256}{conv,\ 3\times3,\hspace{.2cm} 256}\right]\times3$ }& {\small $\left[\bfrac{conv,\ 3\times3,\hspace{.2cm} 256}{conv,\ 3\times3,\hspace{.2cm} 256}\right]\times7$}\\[0.1cm]                                                         
	
	\hline  
	
	{\scriptsize $1\times1$} & \multicolumn{2}{|c|}{{\scriptsize $global\ avg\ pool\ 9\times9$, 10-d \textit{fc}, $softmax$} }\\
	
	\hline                     
\end{tabular}
\caption{{\small Network Structure}}
\label{Table_CIFAR10_NetworkStructure} 
\end{table}

%=================================================================

In the first part of the experiment, both training and inference were performed in the standard way, where training batches were created randomly, and inference were carried out on individual images using a fixed set of means and variances computed using the entire training data after training was completed. In the second part of the experiment, training was done using balanced batches, and inference was carried out twice: First, inference was done in the standard way, where images are tested individually using fixed means and variances. Second, to measure the effect of batch structure on the performance of the network, test images were arranged the same way as training images; test images were arranged as balanced batches, and the means and variances of the current test batch itself were used in the inference process. Table (\ref{Table_CIFAR10_Results}) shows the results for both experiments. It is clear that balancing the training batches doesn't change the results, if the inference process is carried out in a standard way. However, if the performance of the network trained using balanced batches is also tested using balanced test batches, the error rate is reduced by about 80\% for both network models. The error rate was almost eliminated for the non-trivial CIFAR10.  

%==========================================================
\begin{table}[H]
\centering
\renewcommand{\arraystretch}{1.6}
\begin{tabular}{|c|c|c|c|}	
	\hline 
	
	{\scriptsize Training} & {\scriptsize Inference} & {\scriptsize 20 Layers} & {\scriptsize 44 Layers} \\ 
	
	\hline 
	
	{\scriptsize standard} & {\scriptsize standard} & 4.45\% & 3.89\%\\
	
	\hline 
	
	{\small $\mathrm{\bfrac{Balanced}{Batches}}$} & {\scriptsize standard} & 4.47\% & 3.9\%\\                  
	 
	\hline
	
	{\small $\mathrm{\bfrac{Balanced}{Batches}}$} & {\small $\mathrm{\bfrac{Balanced}{Batches}}$} & 0.97\% & 0.69\% \\

	\hline
                    
\end{tabular} 
\caption{{\scriptsize CIFAR10 Results, when training is done using random vs balanced batches, and inference is done on individual images vs balanced batches.}}
\label{Table_CIFAR10_Results}
\end{table}

%==========================================================

\begin{figure}[H]
\centering
\includegraphics[scale=0.8]{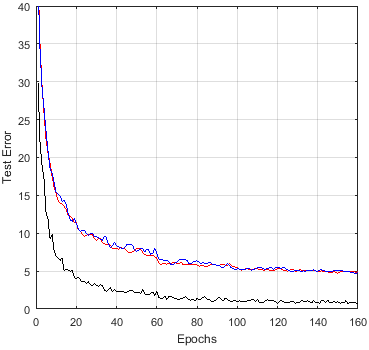} 
\caption{{\scriptsize  validation (test) error curves as training progresses for the CIFAR10 dataset. \textbf{blue curve:} standard training and inference, \textbf{red curve:} training using balanced batches, \textbf{black curve:} both training and inference using balanced batches.}}
\label{CIRFAR10_Curves}
\end{figure}

Figure (\ref{CIRFAR10_Curves}) shows the error curves measured on the central crop of the test set images as training progresses for the 44-layer network. The red curve was obtained by training the network on balanced batches, and the blue curve was obtained by training the network on randomly constructed batches, but because both were measured on individual test images using the running averages of the means and variances, the results were very similar. However, if the progress of the network trained using balanced batches was also measured on balanced test batches using the current means and variances of the batch itself, then a big reduction in the test error can be measured from the start to the end of the training process as the black curve shows. These three curves agree with the final results shown in the three rows in table (\ref{Table_CIFAR10_Results}).

%=================================================================

%================================================================

\subsection{Experiment Two}

Three datasets with 10, 50, and 100 classes were randomly sampled from ImageNet, and will be used here. The reason for not using the entire ImageNet dataset is because it has 1000 classes, and using balanced batches will require a batch size of 1000 image, which cannot be supported by our hardware. The reason for sampling three datasets instead of just one is to measure the impact of batch size on the results obtained using balanced batches. All sampled classes have 1300 training images, and 50 test images. Table (\ref{Table_10_50_100_NetworkStructure}) shows the structure of a 34-layer deep residual network used in this experiment for all three datasets, which is similar to that used by \cite{he2016deep} for ImageNet. Data augmentation in \cite{szegedy2015going}, weight initialization in \cite{he2015delving}, and color augmentation in \cite{krizhevsky2012imagenet} were used. The RMSProp optimization method is used instead of gradient decent with momentum to update the network parameters, using a decay value of 0.999 to calculate the running average per parameter.     

%=================================================================

\begin{figure*}
\centering
\includegraphics[scale=0.6]{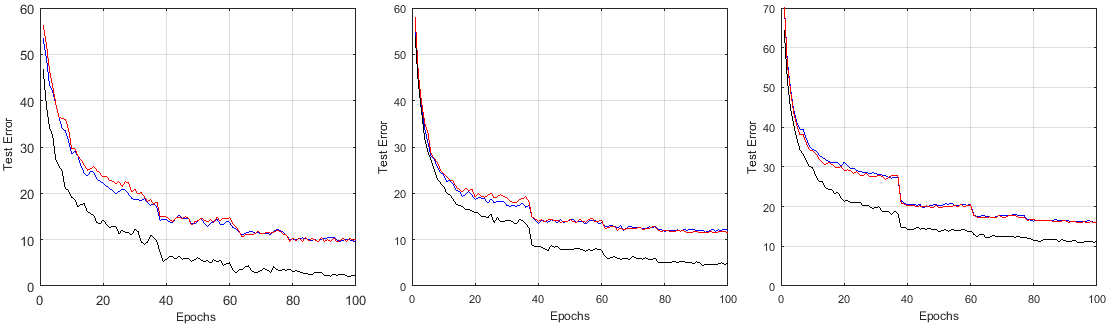} 
\caption{{\scriptsize the validation (test) error curves as training progresses, blue curve using standard training and inference, red curve training using balanced batches, black curve both training and inference using balanced batches. \textbf{left:} for the 10-classes dataset, \textbf{middle:} for the 50-classes dataset, \textbf{right:} for the 100-classes dataset.}   }
\label{10_50_100_Curves}
\end{figure*}

%==========================================================

\begin{table}[H]
\centering
\setlength{\tabcolsep}{3pt}
\renewcommand{\arraystretch}{1.6}
\begin{tabular}{|c|c|c|c|c|}	
	\hline 
	
	{\scriptsize Training} & {\scriptsize Inference} & {\small $\mathrm{\bfrac{10\ Classes}{dataset}}$} & {\small $\mathrm{\bfrac{50\ Classes}{dataset}}$} & {\small $\mathrm{\bfrac{100\ Classes}{dataset}}$} \\ 
	
	\hline 
	
	{\scriptsize standard} & {\scriptsize standard} & 5.97\% & 7.3\% & 10.1\%\\
	
	\hline 
	
	{\small $\mathrm{\bfrac{Balanced}{Batches}}$} & {\scriptsize standard} & 5.9\% & 7.34\% & 10.14\%\\                  
	 
	\hline
	
	{\small $\mathrm{\bfrac{Balanced}{Batches}}$} & {\small $\mathrm{\bfrac{Balanced}{Batches}}$} & 1.1\% & 3.32\% & 6.74\%\\

	\hline
                    
\end{tabular} 
\caption{{\scriptsize results for 3 datasets, when training is done using random vs balanced batches, and inference is done on individual images vs balanced batches.}}
\label{Table_10_50_100_results}
\end{table}

%================================================================

\begin{table}[H]
\centering
\renewcommand{\arraystretch}{1.6}
\begin{tabular}{|c|c|}	
	\hline 
	
	$\mathrm{\bfrac{output}{size}}$ & 34 Layers \\ 
	
	\hline 
	
	{\scriptsize $112\times112$} & {\footnotesize $conv,\ 5\times5,\hspace{.2cm} 96$} \\ 
	
	\hline 
	
	\multirow{2}{*}{\scriptsize {$56\times56$}} & {\footnotesize $max\ pool\ 3\times3$, stride 2}\\
	\cline{2-2}
	                              & {\small $\left[\bfrac{conv,\ 3\times3,\hspace{.2cm} 96}{conv,\ 3\times3,\hspace{.2cm} 96}\right]\times3$} \\[0.1cm]                             
	 
	\hline
	
	\multirow{2}{*}{{\scriptsize $28\times28$}} & {\footnotesize $max\ pool\ 3\times3$, stride 2}\\
	\cline{2-2}
	                              & {\small $\left[\bfrac{conv,\ 3\times3,\hspace{.2cm} 192}{conv,\ 3\times3,\hspace{.2cm} 192}\right]\times4$} \\[0.1cm]                             
	 
	\hline
	
	\multirow{2}{*}{{\scriptsize $14\times14$}} & {\footnotesize $max\ pool\ 3\times3$, stride 2}\\
	\cline{2-2}
	                              & {\small $\left[\bfrac{conv,\ 3\times3,\hspace{.2cm} 384}{conv,\ 3\times3,\hspace{.2cm} 384}\right]\times6$} \\[0.1cm]                             
	 
	\hline
	
	\multirow{2}{*}{{\scriptsize $7\times7$}} & {\footnotesize $max\ pool\ 3\times3$, stride 2}\\
	\cline{2-2}
	                              & {\small $\left[\bfrac{conv,\ 3\times3,\hspace{.2cm} 768}{conv,\ 3\times3,\hspace{.2cm} 768}\right]\times3$} \\[0.1cm]                             
	 
	\hline
	 	
	{\scriptsize $1\times1$} & {\small $\bfrac{global\ avg\ pool\ 9\times9}{10-\mathrm{d}\ fc,\ softmax}$}  \\[0.1cm]
	
	\hline                     
\end{tabular} 
\caption{{\small Network Structure}}
\label{Table_10_50_100_NetworkStructure}
\end{table}

%==========================================================

Table (\ref{Table_10_50_100_results}) shows the results for the three datasets, and compares the results when training and inference are done in the standard way versus using balanced batches. As with the CIFAR10 dataset, training the network using balanced batches doesn't change the results if inference is carried out on individual images using fixed means and variances. However, if the network trained on balanced batches, is also tested using balanced test batches, the error rate is reduced considerably. When using balanced batches, the batch size is equal to the number of classes, and therefore, the batch sizes used here are 10, 50, and 100 images, for datasets with 10, 50, and 100 classes respectively. From table (\ref{Table_10_50_100_results}) The relative reduction in the error when training and inference were done using balanced batches, is very dependent on the batch size. For the 10-classes dataset the reduction was 81\%, for the 50-classes dataset it was 54\%, and for the 100-classes dataset it was 33\%.  This is not surprising because balancing the test batches means that the real task here is not really classifying the images but rather identifying the identity of each image in a balanced batch. And it makes sense that this task gets harder as the size of the batch gets bigger, and as the batch size (thus the number of classes) goes to infinity (or a large number) the performance measured using balanced batches converges to the performance measured on individual images. It is interesting to notice that the 81\% conditional gains for the 10-classes dataset is very close to the 80\% conditional gains obtained with the CIFAR10 data sets, which also has 10 classes. 

Figure (\ref{10_50_100_Curves}) shows the error curves measured on the central crop for all three datasets as training progresses. Again, the red and blue curves show that training the network on balanced batches will produce similar results to training it on randomly constructed batches if the performance is measured on individual test images using the running averages of the means and variances. The black curves show the gains when both training and inference are done on balanced batches using the batch's own means and variances. These learning curves agree with the final results shown in the three rows of table (\ref{Table_10_50_100_results}).

%=================================================================

\begin{figure*}
\centering
\includegraphics[scale=1.0]{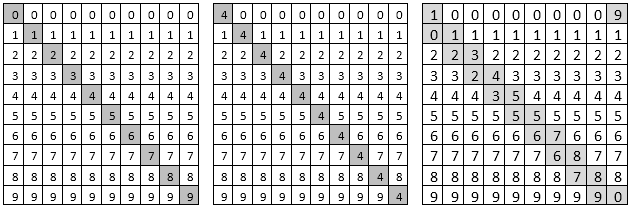} 
\caption{{\scriptsize a balanced test batch made of images classified with high confidence, \textbf{left:} a misclassified test image made to circulate through the batch, \textbf{middle:} correctly classified test image (belong to class no. 4) made to circulate through the batch, \textbf{right:} 2 misclassified test images made to circulate through the batch. } }
\label{10_50_100_Circulate}
\end{figure*}

%=================================================================

\subsection{Shuffling the balanced test batches }

In the training phase shuffling the training images improves the results if BN is implemented in the network. The results also show that if balanced test batches are shuffled, the results will improve even further. It seems that there is an advantage in presenting the different crops of a test image with different batches rather than presenting all of them with the same batch. One possible explanation is that if images are not shuffled, the current balanced batch may contain multiple similar images that belong to different classes, and that may confuse the network. By shuffling the images at inference time, the chances of presenting all the crops of an image with another similar image from a different class will be reduced. Table (\ref{Table_shuffing_results}) shows the error rate measured on shuffled and balanced test batches for the three datasets sampled from ImageNet.  

%==========================================================
\begin{table}[H]
\centering
\setlength{\tabcolsep}{3pt}
\renewcommand{\arraystretch}{1.6}
\begin{tabular}{|c|c|c|c|}	
	\hline 
	
	{\scriptsize Training \& Inference} & {\small $\mathrm{\bfrac{10\ Classes}{dataset}}$} & {\small $\mathrm{\bfrac{50\ Classes}{dataset}}$} & {\small $\mathrm{\bfrac{100\ Classes}{dataset}}$} \\ 
		 
	\hline
	
	{\small $\mathrm{\bfrac{Balanced}{Batches}}$} & 0.2\% & 2.04\% & 5.5\%\\

	\hline
                    
\end{tabular} 
\caption{{\scriptsize Results obtained by shuffling balanced test batches}}
\label{Table_shuffing_results}
\end{table}

%=================================================================

\section{Inference using the batch's own means and variances}

The results in the previous sections show that if batches are structured as balanced batches at the training and inference stages, then the results improve considerably. The experiment presented here shows how the network uses the structure of a balanced test batch in the classification process. This experiment is done on the CIFAR10 dataset using a single balanced test batch made of images identified with high confidence by the network. In the first run a misclassified image (using standard inference) is selected and used to circulate through the created test batch by replacing one image at a time. In the second run an image that has been classified by the network with high confidence (using standard inference) is made to circulate through the batch. Because the number of classes for the CIFAR10 dataset is 10, then each time the selected image will replace one of the 10 images that make up the balanced batch, and the classification of all the images in the batch will be reported. Figure (\ref{10_50_100_Circulate}) shows the classification results, where each column shows a single step, and represents a single batch.

Figure (\ref{10_50_100_Circulate}, left) shows what happened to the weak (misclassified) image, where each time the network classifies the image to belong to the missing class. Figure (\ref{10_50_100_Circulate}, middle) shows the results for the strong (classified with high confidence) image, where the image was always classified correctly, and the identity of the image wasn't changed each time to match the identity of the replaced image. The third run is a generalization of the first one, where two weak images are made to circulate through the balanced batch, replacing two images at a time, and figure (\ref{10_50_100_Circulate}, right) shows how the network has interpreted their identity to replace the missing classes (or one of them).

The results show that when the network is confident about the identity of the test image, then it ignores the structure of the batch. On the other hand, if the network is not confident about the identity of the image, then it relies mainly on the structure of the batch to classify that image. Based on the results shown in figure (\ref{10_50_100_Circulate}), the network uses the structure of a balanced batch as follows: if a balanced test batch made up of $n$ images is passed through the network, and the network is confident of the identity of $m$ out of the total $n$ images, then the network will restrict the prediction of the remaining $n-m$ images to the remaining $n-m$ classes.

%=================================================================

\begin{figure*}
\centering
\includegraphics[scale=0.7]{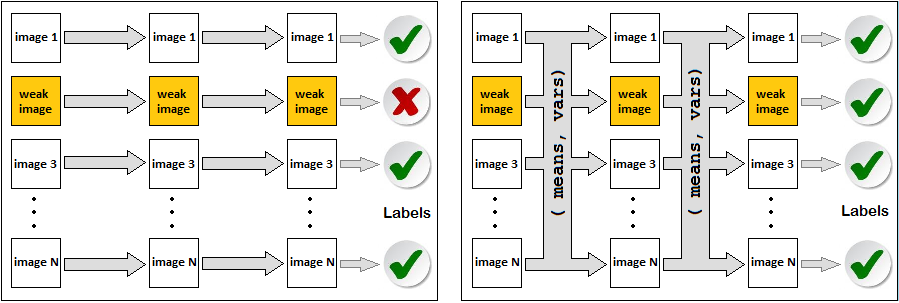} 
\caption{{\scriptsize inference for a balanced batch, \textbf{left:} with precomputed fixed means and variances, \textbf{right:} with current means and variances.}}
\label{shared_means_vars}
\end{figure*}

%=================================================================

%\vspace{1cm}
%\begin{@twocolumnfalse}
%\begin{changemargin}{0cm}{1cm} 
%\begin{minipage}{1.0\textwidth}
%\begin{flushleft}
\begin{mdframed}[hidealllines=true,backgroundcolor=blue!20]
\noindent{\small (\textit{\textbf{m} out of \textbf{n} images identified}) \&\& (\textit{batch is balanced}) $\Rightarrow$ (\textit{remaining images belong to remaining \textbf{n}-\textbf{m} classes})}
\end{mdframed}
%\end{flushleft}
%\end{minipage}
%\end{changemargin} 
%\end{@twocolumnfalse}

Figure (\ref{10_50_100_Circulate}, left) showed how the network was only confident of the identity of 9 out of the 10 images, and therefore the tenth image was always interpreted as the missing class. Figure (\ref{10_50_100_Circulate}, right) showed how the network was only confident of the identity of 8 out of the 10 images, and therefore the identity of the 2 remaining images were always restricted to the 2 missing classes. This logic (highlighted in blue) cannot be implemented by the network if the prediction of one image is independent from all the other images in the batch, and therefore it cannot be implemented without BN. The network was able to implement this logic because it was only exposed to balanced batches in the training stage, and because BN gave the network the capacity to learn something based on the structure of the training batches by using the shared means and variances as a communication tool between the images in the batch.
 
Figure (\ref{shared_means_vars}) explains the process of making a decision about a weak image in a balanced batch, where the network snoops on the decisions about the other images in the batch through the shared means and variances to help decide on the weak image. What is interesting is that this process happens in one forward pass, where the network identifies all images in the current batch at the same time. As signals travel forward, the network found a way to use the shared means and variances at each layer to help guide the prediction of all the images in the batch, based on the fact that batches are balanced. The results presented here show that with BN, controlling the structure of the training batches, adds another layer of control over what the network can learn.

\section{Difficulty balancing the test batches}

\begin{figure}[H]
\centering
\includegraphics[scale=0.7]{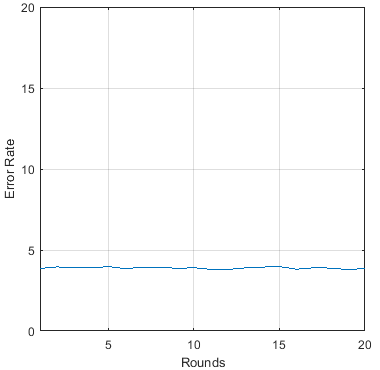} 
\caption{{\scriptsize the repeated process of balancing test batches starting from labels generated by standard inference fails.}}
\label{Circulate_Curves}
\end{figure}

It is tempting to try to translate these big conditional gains into actual gains, however these results are very hard to achieve in practice because structuring test images as balanced batches requires the test image labels. One attempt is to use the labels generated by standard inference to balance the test batches, and then use these semi-balanced batches to generate more accurate labels, where the process is repeated until no improvements can be achieved. Figure (\ref{Circulate_Curves}) shows this process repeated 20 times for the CIFAR10 dataset, but unfortunately it didn't lead to more accurate results (better labels). The error rate stayed almost constant as a horizontal line. From table (\ref{Table_CIFAR10_Results}) the error rate using standard inference is 3.89\%, which means 3.89\% of the images will be misclassified, and those are the toughest images where inference using fully balanced batches made gains to reduce the error to 0.69\%.  From the previous section, figure (\ref{10_50_100_Circulate}) shows what happens to misclassified images when placed in semi-balanced batches, the network often misinterprets their identity to replace the missing class. Therefore, standard inference is not adequate to solve the problem of balancing the test batches, because it will misclassify the important images, where gains need to be made. 

%=================================================================

%=================================================================

\nocite{*}
\bibliographystyle{apalike}
\bibliography{References}
\end{document}